\documentclass[letterpaper, 10 pt, conference]{ieeeconf}  

\IEEEoverridecommandlockouts                              

\usepackage{multirow}
\usepackage{textcomp}
\usepackage{graphics} 
\usepackage{graphicx}
\usepackage{soul}
\graphicspath{{figures/}}
\usepackage[nomessages]{fp}
\makeatletter
\newcommand\newtag[2]{#1\def\@currentlabel{#1}\label{#2}}
\makeatother

\usepackage{etoolbox}
\makeatletter
\patchcmd{\@makecaption}
  {\scshape}
  {}
  {}
  {}
\makeatother

\title{\LARGE \bf
On the Robustness of Speech Emotion Recognition for Human-Robot Interaction with Deep Neural Networks
}

\author{Egor Lakomkin$^{1}$, Mohammad Ali Zamani$^{1}$, Cornelius Weber$^{1}$, Sven Magg$^{1}$ and Stefan Wermter$^{1}$
\thanks{$^{1}$University of Hamburg, Department of Informatics, Knowledge Technology Institute. 
Vogt-Koelln-Strasse 30, 22527 Hamburg, Germany
        {\tt\small $\{$lakomkin, zamani, weber, magg, wermter$\}$ @informatik.uni-hamburg.de }}%
}

\begin{document}

\maketitle
\thispagestyle{empty}
\pagestyle{empty}

\begin{abstract}

Speech emotion recognition (SER) is an important aspect of effective human-robot collaboration and received a lot of attention from the research community. For example, many neural network-based architectures were proposed recently and pushed the performance to a new level. However, the applicability of such neural SER models trained only on in-domain data to noisy conditions is currently under-researched. In this work, we evaluate the robustness of state-of-the-art neural acoustic emotion recognition models in human-robot interaction scenarios. We hypothesize that a robot's ego noise, room conditions, and various acoustic events that can occur in a home environment can significantly affect the performance of a model. We conduct several experiments on the iCub robot platform and propose several novel ways to reduce the gap between the model's performance during training and testing in real-world conditions. Furthermore, we observe large improvements in the model performance on the robot and demonstrate the necessity of introducing several data augmentation techniques like overlaying background noise and loudness variations to improve the robustness of the neural approaches.

\end{abstract}


\section{INTRODUCTION}

\par Emotions and, in general, affective state recognition play an important role in communication between humans and they allow us to evaluate intentions and urgency of a message quickly. Emotions can be expressed very differently given the speaker's cultural background, context and environmental conditions and, as a result, models that learn from unstructured data are very effective in this case. Complex pattern recognition models like deep neural models led to many recent advances in solving difficult problems like speech recognition \cite{Amodei2016DeepMandarin}, image classification \cite{Krizhevsky2012ImageNetNetworks} or robot motion planning \cite{Finn2017Deep}. 

 \begin{figure}[h]
  \includegraphics[width=\linewidth]{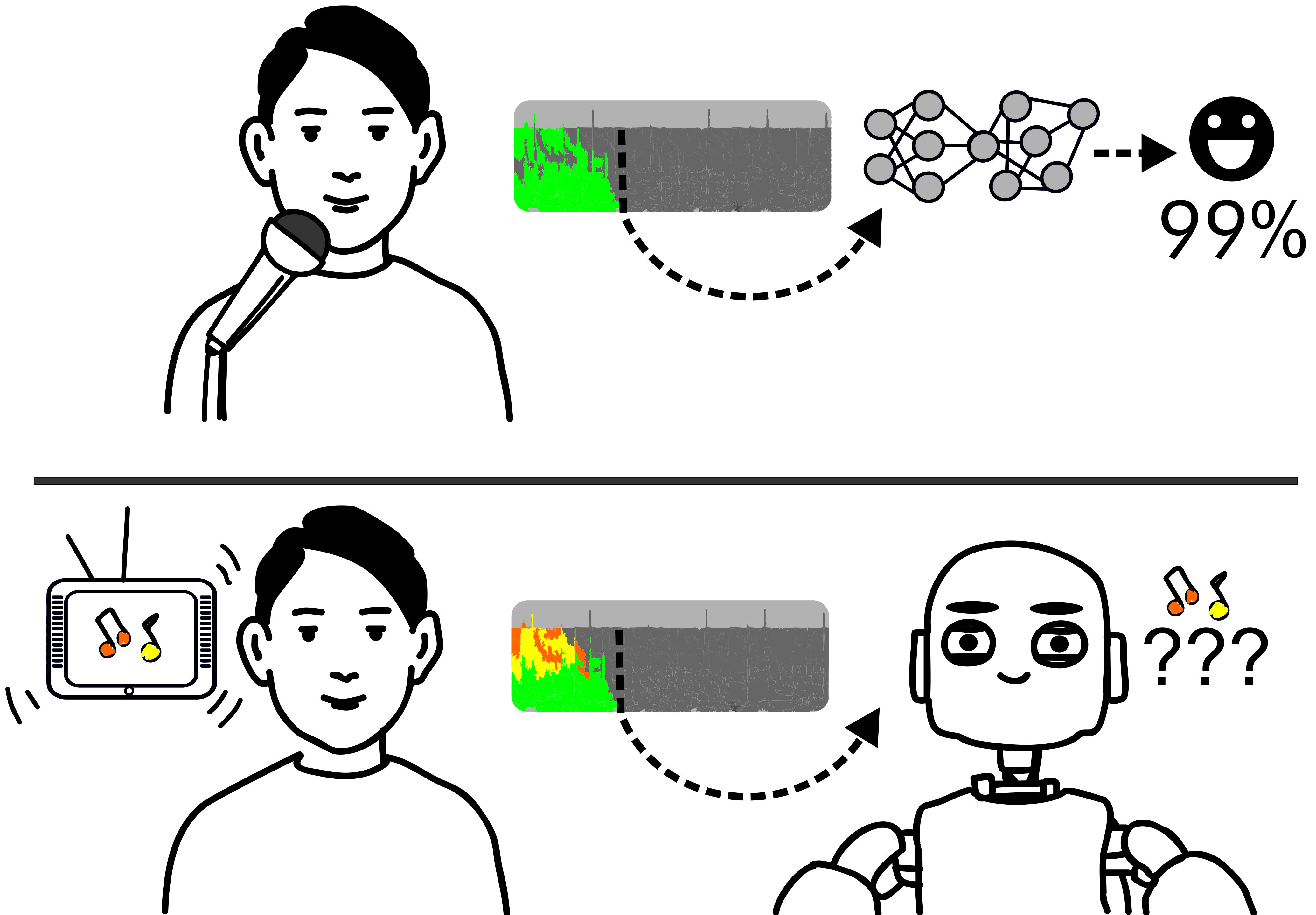}
  \caption{ Illustration of the core problem that we address in the paper. Neural speech emotion recognition models are commonly trained and evaluated on in-domain data collected in a constrained environment. When deployed on a robot, these models can expect a performance degradation due to the environmental conditions and presence of the robot's ego-noise.  }
  \label{fig:system_main}
\end{figure}

An essential ingredient to train neural networks is the training data. One common problem that affects the performance of the machine learning model is overfitting, when a model with high capacity like neural networks can fit the training data very well while performing poorly on the unseen data. Several regularization techniques were proposed to mitigate overfitting like dropout \cite{srivastava2014dropout}, batch-normalization \cite{ioffe2015batch, Cooijmans2017RecurrentNormalization} and layer normalization \cite{Ba2017LayerNormalization}. Another possible solution is data augmentation which introduces deformations to the input while not changing its label, for example, by varying the tempo or pitch of a spoken utterance. For example, Tarvainen and Valpola  demonstrated that enforcing consistency between the prediction of the original and the corrupted sample greatly improves the performance and robustness of the model \cite{Tarvainen2017MeanResults}.

\par The problem of discrepancy between training and testing conditions is especially relevant for robotic applications. For example, different types of microphones can be mounted on a robot or the robot can be present in very different environmental conditions, like small rooms or larger halls. The source of the sound can also have  an arbitrary distance to the robot, which affects the signal-to-noise ratio. In addition, there are several sources of ego noise coming from fans, hardware, and joint movements, which complicate speech and emotion recognition. Also, robots will be naturally present in homes or offices where the important speech signal can be masked by various external noise and audio events. We hypothesize that noise can greatly affect the model's performance \cite{Abdelwahab2015SupervisedSpeech}, especially, since many emotion-labelled datasets are collected in the controlled condition of a lab environment. In our work, we evaluate the performance of state-of-the-art neural network models on acoustic emotion recognition tasks in a realistic noise environment and perform the experiments on the iCub robot.

\par Our contribution is two-fold: a) We evaluate the performance of state-of-the-art neural acoustic emotion recognition models in a set-up that simulate real-world scenarios, like recognizing emotion of a human when the sound is overlaid with noise (e.g. robot's ego noise). b) We propose and evaluate several speech data augmentation techniques and analyze their effects on the performance of the neural models in acoustically clean conditions and when recorded by the iCub.

\par The paper is organized as follows: section II introduces related work and section III describes the methodology including feature extraction from the acoustic signal and data augmentation steps to improve the robustness of the model. We describe several neural models used in our experiments such as recurrent and convolutional neural networks. Section IV outlines the conducted experiments on the iCub robot head.

\par We found that the models that are trained on clean data, can fail when they are deployed on the robot due to reasons like ego noise or room conditions. However, we could reduce this performance drop using different data augmentations such as changing tempo and loudness, adding Gaussian and background noise, and convolving signals with different room impulse response functions. Moreover, we observed a significant increase in the performance on the iCub-recorded data when we included data and noise augmentations in our training pipeline. We achieved these improvements on the iCub neither using any training data samples from the robot (e.g. recording robot's ego-noise) nor the real lab impulse responses during training and validation. This result shows the importance of adding data augmentations during training for the SER task to make a model robot independent and more robust in general.

\section{RELATED WORK}
\begin{figure*}[t]
\vspace{5pt}
  \includegraphics[height=.43\linewidth]{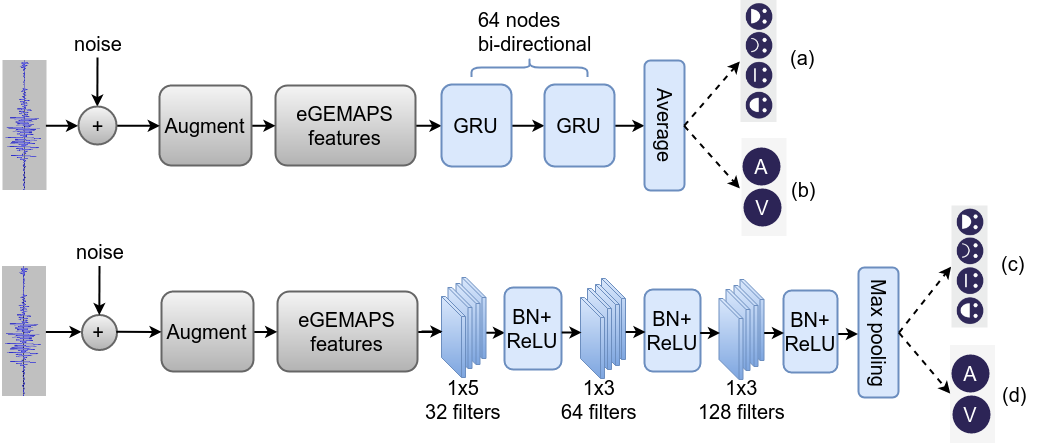}
  \caption{A recurrent neural network (RNN) model, containing two bi-directional GRU layers, followed by a temporal averaging layer (e.g. uniform attention) for categorical (a) and dimensional cases (b). A convolutional neural network (CNN) model, containing three 1-dimensional convolution layers, followed by a batch-normalization layer and ReLU activation function for categorical (c) and dimensional (d) cases. Both architectures are trained with the same feature set: extended GEMAPS, optionally performing random data augmentation and noise injection before the feature extraction step.}
  \label{fig:rnn-model}
\end{figure*}

\par Deep neural networks significantly boosted the performance of acoustic emotion recognition models. The majority of recent work focuses on learning to extract useful input representations and searching for neural architectures for emotion recognition, as neural approaches outperform traditional ones like support vector machines and decision trees \cite{Fayek2017EvaluatingRecognition}.  

\par Recurrent neural networks have an ability to model long-term context information and were successfully applied to emotion recognition \cite{Huang2016AttentionRecognition, Lee2015High-levelRecognition}. Convolutional neural networks can capture only a local context, but have an ability to model longer dependencies when their architecture was designed with a deep hierarchy \cite{Fayek2017EvaluatingRecognition}. Commonly, these methods train neural networks on pre-extracted features: MFCC coefficients, spectrograms and high-level information like formants, pitch, and voice probability. Alternatively, Trigeorgis et al. demonstrate a model that learns how to recognize the affective state of a person directly from the raw waveform  \cite{Trigeorgis2016AdieuNetwork}. Another explored direction is transfer learning: adapting audio representations trained initially for other auxiliary tasks, like gender and speaker identification \cite{Gideon2017ProgressiveRecognition} or speech recognition \cite{Fayek2016OnRecognition, Lakomkin2017ReusingRecognition}. 

\par Robustness to noise was a subject of several previous work. Attention mechanisms \cite{Huang2016AttentionRecognition, Neumann2017AttentiveSpeech} aim to identify useful regions for emotion classification automatically by assigning a low importance to irrelevant inputs, for example, non-speech or silence frames. Adding background noise during training improved the robustness of neural models in several acoustic classification tasks \cite{Lane2015DeepEar:Learning}. Different types of data augmentation methods were explored by Zhou et al. \cite{Zhou2017ImprovedRecognition} to improve the performance of speech recognition. Supervised domain adaptation was proposed by Abdelwahab et al. \cite{Abdelwahab2015SupervisedSpeech} to mitigate the problem of training and testing mismatch conditions by tuning the model on the small set of test samples.

\par Our work is close to Lane et al. \cite{Lane2015DeepEar:Learning} and our main difference is that our testing conditions are not synthetically constructed by overlaying clean samples with additive noise, but recorded on the iCub robot which adds a significant amount of ego-noise. We argue that distortions introduced by playing a sample through speakers, changing room conditions and distance from the speech source to the robot, reverberations, added external acoustic events and the robot's internal noise introduce non-linear deformations which are challenging for the neural network to deal with.

\section{METHODOLOGY}
In this section, we describe a feature extraction procedure from the acoustic signal and outline several proposed data augmentation methods. Then, we introduce the neural architectures that we used in our experiments and the training details. 
\subsection{Feature Extraction}

We extracted 32 low-level features from the eGEMAPS low-level descriptors \cite{Eyben2016TheComputing} using the OpenSMILE\footnote{https://audeering.com/technology/opensmile/}\cite{Eyben2013RecentExtractor} toolkit. It contains frequency-related features (pitch, jitter, formant information), energy-related features (shimmer, loudness), and spectrum-related features (13 mel-frequency cepstrum coefficients, spectral flux) extracted from a 25ms window with 10ms stride. The original eGEMAPS feature set includes only the first four MFCC coefficients, but in our experiments, we found that using 13 coefficients improves the performance. All coefficients were smoothed with a window size of 3. We calculated mean and standard deviation for each feature value over the whole training set and used them to normalize the samples during training and testing.

\subsection{Data Augmentation}

Previous research in end-to-end speech recognition demonstrated the importance of introducing random perturbations into the speech signal like a change of pitch, tempo, loudness, and adding noise \cite{Amodei2016DeepMandarin, Hannun2014DeepRecognition, Zhou2017ImprovedRecognition}. As such perturbations do not alter the target label (spoken text in the case of speech recognition or an emotion category), they can be conveniently applied with some occurrence probability during training. Data augmentation can be considered also as a way to increase the training data size. We will show that this procedure already reduces overfitting. In the case of acoustic emotion recognition, neural models overfit, since the available labelled data is sparse in terms of the number of samples, the variety of speakers and recording conditions.

\par In our experiments, we 1) changed the tempo of the recording by sampling the speed factor uniformly in a range of [85, 120] percent, 2) changed the loudness of the recording by sampling gain uniformly in a range of [-6, 3] dB, 3) added random background noise (more in section IV.B) by sampling the noise-to-signal ratio uniformly in the range [0.5, 0.9], and 4) applied room filter impulse responses selected randomly from the Aachen Impulse Response database \cite{Jeub2009AAlgorithms}. We used the SoX\footnote{http://sox.sourceforge.net/}  utility to perform all data augmentation steps. 
It is important to note that we did not use any sample of data corresponding to the iCub's ego-noise during training or validation. Our goal was to identify the model's performance in conditions it never observed during training.

\subsection{Emotion Recognition Model}

Neural network-based models recently demonstrated state-of-the-art performance in different affective modelling tasks: multi-modal sentiment analysis \cite{Poria2017Context-DependentVideos}, facial expression \cite{Barsoum2016TrainingDistribution} and emotion recognition \cite{Huang2016AttentionRecognition}. In our experiments, we compared convolutional neural network (CNN) and recurrent neural networks (RNNs), as the two most popular approaches to process variable-length speech sequences.  

\subsection{Recurrent Neural Network Model}

Recurrent neural networks demonstrated state-of-the-art performance in the affective modelling tasks of sentiment analysis \cite{Poria2017Context-DependentVideos} and emotion recognition \cite{Huang2016AttentionRecognition, Lakomkin2017ReusingRecognition}. In our experiments, we evaluated a two-layer bi-directional recurrent neural network with Gated Recurrent Unit (GRU) \cite{Bahdanau2015NeuralTranslate} followed by a softmax layer with four output nodes, in the categorical case, modelling a distribution over four emotion classes, which is overall similar to the model used in Huang et al. \cite{Huang2016AttentionRecognition}. In the dimensional case, the GRU is followed by two linear nodes for arousal and valence. We use the same hyperparameters and training set-up as presented in Huang et al. \cite{Huang2016AttentionRecognition} in all our experiments (\textit{RNN} in the results table \ref{all-results}), and the scheme of the model is presented in Fig. \ref{fig:rnn-model}. 

\subsection{Convolutional Neural Network Model}

Convolutional neural networks were successfully applied to acoustic emotion recognition problems \cite{Neumann2017AttentiveSpeech} and as demonstrated in Fayek et al. \cite{Fayek2017EvaluatingRecognition} can show competitive results compared with RNNs. Though CNNs, as opposed to RNNs, do no have an explicit memory vector to retain useful contextual information while processing the whole sequence, they are capable of modelling long sequences by using progressively large receptive fields by stacking several CNN layers on top of each other \cite{Kim2014ConvolutionalClassification} and using various types of pooling to encode information in the sequence \cite{Neumann2017AttentiveSpeech}.  We used a 3-layer convolutional neural network architecture with 1-dimensional filters (see Fig. \ref{fig:rnn-model}), where each convolution layer is followed by the batch normalization layer and ReLU as the non-linear activation function. Max temporal pooling was applied to encode the sequence into a fixed-length vector.
\subsection{Training details}

We used the Adam optimizer \cite{Kingma2014Adam:Optimization} with a learning rate of 3e-4, clipped the gradient values to keep them in the interval [-1, 1] and used a batch size of 32. If the results on the validation set did not improve over the course of training, we reduced the learning rate by a factor of 2. In addition, we followed the SortaGrad \cite{Amodei2016DeepMandarin} training routine by presenting samples to the network in a sorted way during the first epoch.

\section{EXPERIMENTAL RESULTS}
We conducted several experiments to evaluate the impact of data augmentation on the performance of the proposed methods on the iCub robot. In this section, we describe the dataset that we used for training, the evaluation procedure, and achieved results.

\begin{table*}[]
\centering 
\caption{Evaluation results. We evaluated two neural network architectures (the RNN and CNN), trained on the IEMOCAP dataset and with data augmentation (IEMOCAP + augmentation and noise) and tested on the IEMOCAP original samples and re-recorded on the iCub (\textit{IEMOCAP-iCub}). Metrics reported: unweighted accuracy, unweighted average recall, macro F-score, arousal and valence mean absolute error, and Pearson's correlation coefficient. Also, we reported a gap in performance for each model evaluated in clean and noisy conditions (higher is better for UW Acc, UAR, F-score, Arousal and valence corr and lower is better for Arousal and Valence MAE) and relative performance improvement on the \textit{IEMOCAP-iCub} by adding augmentations during training.}
\label{all-results}
\begin{tabular}{ccc|ccc|cccl|}
\cline{4-10}
\multicolumn{1}{l}{}        & \multicolumn{1}{l}{}                                                                                & \multicolumn{1}{l|}{} & \multicolumn{3}{c|}{Categorical}                             & \multicolumn{4}{c|}{Dimensional}                                               \\ \hline
\multicolumn{1}{|c|}{Model} & \multicolumn{1}{c|}{Train conditions}                                                               & Test conditions       & UW Acc             & UAR                & F-score            & \begin{tabular}[c]{@{}c@{}}Arousal\\ MAE\end{tabular}         & \begin{tabular}[c]{@{}c@{}}Valence\\ MAE\end{tabular}  & \begin{tabular}[c]{@{}c@{}}Arousal\\ corr\end{tabular}  & \begin{tabular}[c]{@{}c@{}}Valence\\ corr\end{tabular}     \\ \hline

\multicolumn{1}{|c|}{}      & \multicolumn{1}{c|}{\multirow{3}{*}{IEMOCAP}}                                                       & IEMOCAP             & 0.533              & 0.559              &  \newtag{0.531}{RNN_cln_trn_cln_test}             & 0.430            & 0.655             & 0.715             & 0.525              \\   
\multicolumn{1}{|c|}{}      & \multicolumn{1}{c|}{}                                                                               & \textit{IEMOCAP-iCub}                  & 0.203              & 0.303              & \newtag{0.144}{RNN_cln_trn_icub_test}       & 0.493           & 1.004             & 0.572            & 0.076            \\ 
\multicolumn{1}{|c|}{RNN}   & \multicolumn{1}{c|}{}                                                                               & Gap \%                & -61.9\%            & -45.8\%            & -72.9\%            & +14.6\%           & +53.2\%           & -20.1\%           & -85.5\%           \\ \cline{2-10} 

\multicolumn{1}{|c|}{}      & \multicolumn{1}{c|}{\multirow{4}{*}{\begin{tabular}[c]{@{}c@{}}IEMOCAP\\ + Augmentation\\ and noise\end{tabular}}} & IEMOCAP               & 0.545              & 0.563              & \newtag{0.54}{RNN_aug_trn_cln_test}             & 0.422             & 0.727             & 0.675             & 0.426              \\ 
\multicolumn{1}{|c|}{}      & \multicolumn{1}{c|}{}                                                                               & \textit{IEMOCAP-iCub}                  & 0.475              & 0.418              & \newtag{0.411 }{RNN_aug_trn_icub_test}              & 0.431             & 0.762             & 0.658             & 0.33              \\ 
\multicolumn{1}{|c|}{}      & \multicolumn{1}{c|}{}                                                                               & Gap \%                & -12.8\% & -25.71\% & -23.9\% & +2.1\% & +4.8\% & -2.5\% & -22.5\% \\ \cline{3-10} 

\multicolumn{1}{|c|}{}      & \multicolumn{1}{c|}{}                                                                               & Improvement \%                & \textbf{+134.0\%} & \textbf{+37.9\%} & \textbf{+185.4\%} & \textbf{+12.5\%} & \textbf{+24.1\%} & \textbf{+15.0\%} & \textbf{+334.2\%} \\ \hline

\multicolumn{1}{|c|}{}      & \multicolumn{1}{c|}{\multirow{3}{*}{IEMOCAP}}                                                       & IEMOCAP               & 0.511              & 0.532                & \newtag{0.505}{CNN_cln_trn_cln_test}            & 1.351                 & 1.150                & 0.687                &     0.412               \\ 
\multicolumn{1}{|c|}{}      & \multicolumn{1}{c|}{}                                                                               & \textit{IEMOCAP-iCub}                  & 0.360              & 0.342                & \newtag{0.247}{CNN_cln_trn_icub_test}            & 1.419                 & 1.116                 & 0.647                 &         0.155           \\ 
\multicolumn{1}{|c|}{CNN}   & \multicolumn{1}{c|}{}                                                                               & Gap                   & -29.5\%              & -35.7\%                & -51.1\%             & +5\%                 & +2.9\%                 & -5.4\%                 &      -62.3\%             \\ \cline{2-10} 
\multicolumn{1}{|c|}{}      & \multicolumn{1}{c|}{\multirow{4}{*}{\begin{tabular}[c]{@{}c@{}}IEMOCAP\\ + Augmentation\\and noise\end{tabular}}} & IEMOCAP               & 0.495              & 0.521                & \newtag{0.48}{CNN_aug_trn_cln_test}             & 1.320                 & 1.184                 & 0.638                 &      0.214              \\ 
\multicolumn{1}{|c|}{}      & \multicolumn{1}{c|}{}                                                                               & \textit{IEMOCAP-iCub}                  & 0.400              & 0.401                & \newtag{0.312}{CNN_aug_trn_icub_test}            & 1.399                 & 1.164                 & 0.605                 &     0.145               \\ 
\multicolumn{1}{|c|}{}      & \multicolumn{1}{c|}{}                                                                               & Gap \%                  & -19.2\%              & -23\%              & -35\%             & +5.9\%                 & -1.6\%                 & -5.1\%                 &         -32.2\%           \\ \cline{3-10} 
\multicolumn{1}{|c|}{}      & \multicolumn{1}{c|}{}                                                                               & Improvement \%                   & \textbf{+11.1\%}              & \textbf{+17.2\%}                & \textbf{+26.3\%}             & \textbf{+1.4\%}                 & -4.3\%                 & -6.49\%                 &         -6.45\%            \\ \hline
\end{tabular}
\vspace{20pt}
\end{table*}

\subsection{Data}


\par We used the IEMOCAP \cite{Busso2008IEMOCAP:Database} dataset for our experiments, which contains five sessions with two actors in each, performing either scripted dialogues or improvising on several pre-defined topics (e.g. unsatisfied customer at the bank or sharing a happy moment with a friend) resulting in 10,030 utterances and 12 hours of speech overall. Each utterance is labelled by three to five annotators with categorical labels (\textit{Angry, Happy, Neutral, Sad, Excited, Fear}, and  \textit{Disgust}) and dimensional values: valence and arousal. Valence represents a value from 1 (very negative) to 5 (very positive). The arousal value reflects the degree of excitement of the person, where 1 is very calm and 5 is very active speech. In our experiments, we only used samples labelled as \textit{Angry, Happy, Neutral}, and \textit{Sad}, as they are the most often occurring ones in the dataset and also to be consistent with previous work in HRI. As in multiple previous work \cite{Neumann2017AttentiveSpeech,Fayek2017EvaluatingRecognition}, we merged the \textit{Happy} and \textit{Excited} classes together. The data distribution was 1,103 \textit{Angry}, 1,708 \textit{Neutral}, 1,636 \textit{Happy} and 1,084 \textit{Sad} samples resulting in 5,531 utterances overall.

\subsection{iCub Data}

To evaluate the model deployed on the robot, we recorded the IEMOCAP dataset in the Knowledge Technology human-robot interaction lab. We played IEMOCAP utterances through the speakers and recorded the signal captured by two microphones on the iCub head (mounted in the left and right ears with realistic pinnae). We do not do any signal processing (like changing loudness) before playing the recording.

\subsubsection{Lab Setup}

\begin{figure}[b]
  \includegraphics[width=\linewidth]{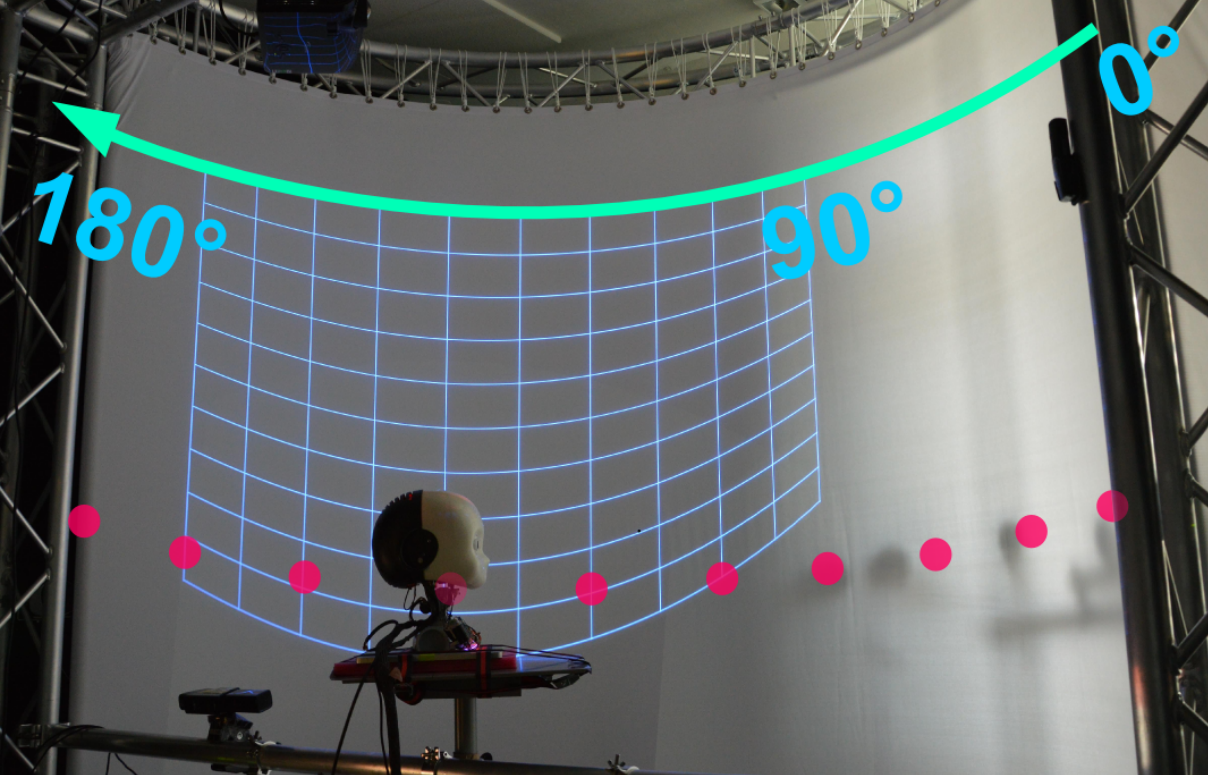}
  \caption{Lab setup of the iCub in front of loudspeakers behind a screen. See also \cite{Bauer2012SmokeRobotics}.}
  \label{fig:lab_setup}
\end{figure}

Our goal is to simulate a scenario close to a real-life situation. The experimental setup which is shown in Fig. \ref{fig:lab_setup}, consists of a humanoid robot head (iCub) immersed in a display to create a virtual-reality environment for the robot  \cite{Bauer2012SmokeRobotics}. Loudspeakers are located behind the display between 0\textdegree \ and 180\textdegree \ along the azimuth plane with the same elevation. The iCub head is 1.6 meters away from the speakers. The setup introduces background noise generated by the projectors, computers, power sources as well as ego noise from the iCub head.

\subsubsection{Background Noise and Acoustic Events} 
As we expect that robots will eventually share home environments with humans, we are interested in evaluating acoustic emotion recognition models in conditions similar to real-life situations. To this end, we play different natural noise samples: air conditioner, babbling noise (TV), and salient loud events like door knock or cell phone ring-tone. When such noise overlays speech, it can introduce distortions that might influence the neural network performance. We utilize two resources to select noise samples: Freesound\footnote{https://freesound.org/} and UrbanSound 8k \cite{Salamon2014AResearch}. Freesound is a collection of audio samples uploaded by users with a short textual description and a list of tags. We fetched samples labelled with tags like \textit{clap, click, crash}, or \textit{kitchen} to collect a set of audio events that can occur at a home environment. In addition, we manually constructed an ``unwanted'' list of tags (like, \textit{sfx} ) and we filtered  samples annotated with any tag from it. The full list of ``desired'' tags is: \textit{mash, break, crash,  accident, shatter,  crack, cracking, kitchen, knock, knocking, domestic-sounds, collapse, alarm, warning, horn, fire-alarm, alert, gunfire, siren, tap, beep, falling,  snapping, household}, and \textit{falling}. The unwanted list is composed of human-produced vocal sound that can interfere with the task (e.g. \textit{speech, voice, cry, scream, shout, pain, crying}, and \textit{cough}), irrelevant (e.g. \textit{nature}, and \textit{field-recording}) and synthesized sounds (e.g. \textit{special-effects, synthesizer}, and \textit{sound-effect}).  

\begin{table}[t]
\centering
\caption{Ablation study on different augmentation techniques. A relative difference in the F-score when one augmentation method is excluded from the training pipeline of the RNN model (base F-scores are \ref{RNN_aug_trn_cln_test} and \ref{RNN_aug_trn_icub_test} for IEMOCAP and \textit{IEMOCAP-iCub}, respectively).}
\label{ablation-augmentation}
\begin{tabular}{ccc}
Augmentation type        & IEMOCAP & \textit{IEMOCAP-iCub}     \\ \hline
-tempo                   & +2.3\%  & -6\%             \\
-loudness                & +5.1\%  & -2.7\%           \\
-background noise        & +9.7\%  & \textbf{-36.8\%} \\
-filter impulse response & +5.3\%  & +1.9\%          
\end{tabular}
\end{table}
\subsection{Experiments}

\par We followed the leave-one-speaker-out cross validation to report the performance of the model in all our experiments. As the IEMOCAP dataset contains five sessions with two speakers in each, we used four sessions for training and the remaining one for testing and validation (samples from one speaker were used for validation and from the other one for testing), resulting in ten folds overall. Original IEMOCAP samples were used when testing the models in clean conditions and re-recorded on the iCub in the \textit{IEMOCAP-iCub} experiments.  Thus we can evaluate the robustness of the models by testing them in two different testing conditions. 

\par As the IEMOCAP dataset provides both categorical and dimensional labels, we also used them both in our experiments, and the models were trained independently for categorical and dimensional labels. We present the results in Table \ref{all-results}. We report unweighted accuracy, unweighted average recall and macro F-score for categorical labels and mean-absolute error and Pearson's correlation coefficient of arousal and valence for dimensional labels.

\par We tested the trained model on two setups: samples from the original IEMOCAP dataset which we refer to as clean data (IEMOCAP in table \ref{all-results}), and samples from the IEMOCAP dataset re-recorded on the iCub (\textit{IEMOCAP-iCub} in table \ref{all-results}).
The sample IDs in all these two sets were identical and taken from the IEMOCAP dataset, but the recording conditions were different as described above. We hypothesize that emotion detection in the \textit{IEMOCAP-iCub} dataset should be a more challenging task, as it contains the robot's ego noise that can significantly corrupt the original sample.

\subsection{Results}
\begin{figure}[t]
  \includegraphics[width=\linewidth]{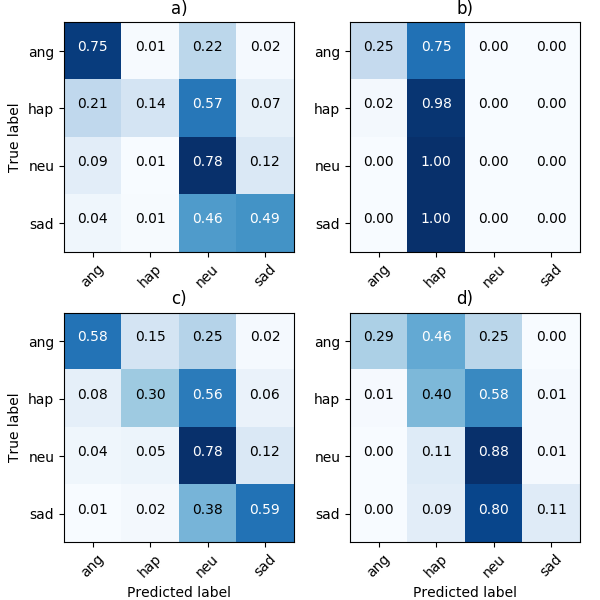}
  \caption{A confusion matrix for our RNN model trained on the original IEMOCAP dataset and evaluated on the IEMOCAP test set (a), \textit{IEMOCAP-iCub} samples (b), trained with data augmentations and noise injection and evaluated on the IEMOCAP dataset (c), and \textit{IEMOCAP-iCub} samples (d). }
  \label{fig:conf-matrix}
\end{figure}

 \begin{figure}[h]
  \includegraphics[width=\linewidth]{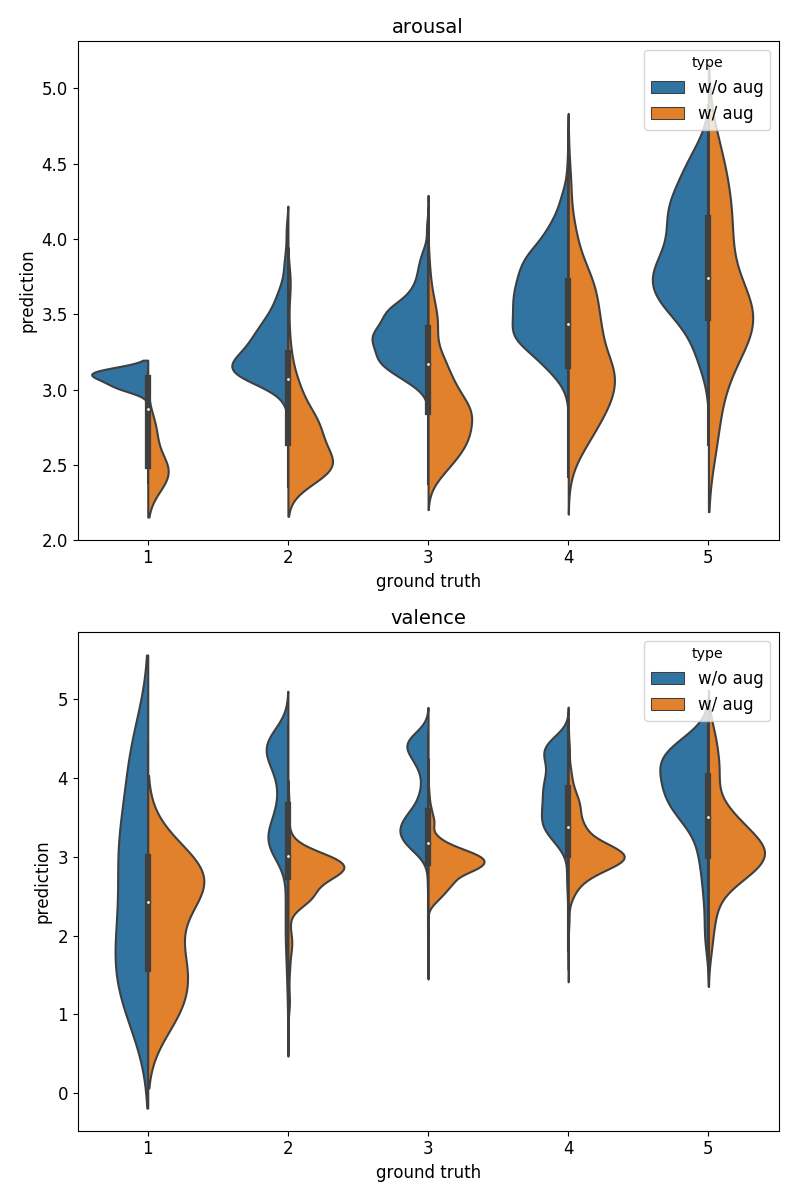}
  \caption{Violin plot of ground-truth arousal (top) and valence (bottom) (x-axis) vs predicted arousal and valence (y-axis) tested on the iCub recordings of our RNN model with no augmentation (blue) and augmentations turned on (orange). The plot shows full distribution of model's predictions given the ground truth value. Adding augmentations leads to more neutral arousal compared to w/o augmentation which predicts too high arousal, which could be explained by that the network has not learnt noise and thus does not ignore acoustic frames containing only noise and can interpret them as high arousal speech. We observe similar behavior for the valence parameter as well.}
  \label{fig:scatter}
\end{figure}

\par We present our results in Table \ref{all-results}. The performance of our RNN model on the IEMOCAP test dataset match previously reported results by Huang et al. \cite{Huang2016AttentionRecognition}. We observe a significant difference in the performance between IEMOCAP and \textit{IEMOCAP-iCub} test conditions and consistent improvements on the \textit{IEMOCAP-iCub} when we add augmentations. This result proves that neural acoustic emotion recognition is very sensitive to the recording conditions. The F-score of the RNN model without any augmentations drops from \ref{RNN_cln_trn_cln_test}  on  IEMOCAP down to \ref{RNN_cln_trn_icub_test} on \textit{IEMOCAP-iCub}. The F-score for the CNN also drops from \ref{CNN_cln_trn_cln_test}  to \ref{CNN_cln_trn_icub_test}. When we add augmentations and additive noise during training, the F-scores on the \textit{iCub} test set rise to \ref{RNN_aug_trn_icub_test} for the RNN and \ref{CNN_aug_trn_icub_test} for the CNN, which are significant improvements.

\par We noted that \textit{Sadness} and \textit{Neutral} samples are particularly difficult to recognize (see Figure \ref{fig:conf-matrix}) and we hypothesize that \textit{Sadness} samples have a lower signal-to-noise ratio when recorded on the iCub, as in the majority of samples belonging to the \textit{Sadness} class an actor is speaking quietly in the IEMOCAP dataset. Adding augmentations mitigates the problems  for the \textit{Neutral} class, but it remains still very difficult to classify \textit{Sadness} samples correctly. 

\par Among the two-dimensional parameters, valence is the most affected one (see Fig. \ref{fig:scatter}) when evaluated on the iCub. The model without augmentations tend to overshoot the value of valence and have higher variance in its predictions compared to the model trained with augmentations. Valence (or degree of negativity/positivity) is naturally very difficult to evaluate given only acoustic signals as it depends also on the content of spoken text. 

In addition, we note that there is no significant difference in the performance on the IEMOCAP test set, when we add noise augmentations during training, which leads us to the conclusion that it is safe to add noise augmentation to the training pipeline to improve the robustness of the neural model without jeopardizing the expected performance on the original data. Compared with the models that are trained on the IEMOCAP dataset, we gained 0.009 and lost 0.025 of F-score for the RNN and CNN, respectively, when we trained with data augmentations and tested on IEMOCAP (clean samples). 

We conducted an ablation study (see Table \ref{ablation-augmentation}) to evaluate effects of different data augmentation techniques on the performance of the clean samples (IEMOCAP data) and the noisy samples (\textit{IEMOCAP-iCub} data). We measured the performance difference when one of the data augmentation methods was excluded compared with the training regime when all of them were included. In our ablation study, we found that adding background noise was the most effective augmentation type and excluding it led to a 36\% F-score drop on the \textit{IEMOCAP-iCub} dataset. Interestingly, varying speech tempo led to the least increase in the performance of the IEMOCAP dataset while it was ranked as the second most important augmentation type according to our ablation study.

\section{CONCLUSIONS}

\par In this paper, we evaluated two neural speech emotion recognition models and showed that they perform significantly worse when trained only on in-domain clean data and tested on the iCub robot. We demonstrated that data augmentation reduced this performance loss and overall significantly improved the robustness of the model even without using any real robot ego noise or room conditions. We observed significant performance gains on the \textit{IEMOCAP-iCub} data by injecting noise during training. We thus can conclude that it is crucial to include training data augmentations to prepare models for deployment on a robot.
\par In future work, we plan to investigate further ways to enhance the data augmentation pipeline. For example, data-driven generative models, like generative adversarial networks, can produce realistic speech samples, which potentially can be useful during training. Also, we plan to evaluate an option to enrich input representation with the information on the spoken text under noisy conditions as it appears  to be difficult to analyze valence without it.  






\section*{ACKNOWLEDGMENT}
The authors thank Erik Strahl for his continuous support with the experimental setup and the iCub and Julia Lakomkina for her help with illustrations.
This project has received funding from the European Union's Horizon 2020 research and innovation programme under the Marie Sklodowska-Curie grant agreement No 642667 (SECURE) and Crossmodal Learning (TRR 169).


\addtolength{\textheight}{-5.7cm}   

\bibliography{Mendeley_egor,maz}

\end{document}